\documentclass[letterpaper, 10 pt, conference]{class/ieeeconf}  % Comment this line out
\IEEEoverridecommandlockouts                              % This command is only
\usepackage{amsmath}
\usepackage{amssymb}
\usepackage{latexsym}
\usepackage{url}
\usepackage{cite}
\usepackage{relsize}
\usepackage{multirow}
\usepackage{afterpage}
\usepackage{ifthen}
\usepackage{graphicx}
\usepackage{algpseudocode,algorithm,algorithmicx}
\usepackage{subfigure}
\usepackage[keeplastbox]{flushend}
\usepackage{epstopdf}
\usepackage{stackengine}
\usepackage{gensymb}
\usepackage[bookmarks=false, linkcolor=blue, urlcolor=blue, citecolor=blue]{hyperref} 
\usepackage{booktabs}
\usepackage{makecell}
\usepackage{hhline}
\usepackage{lipsum}
\usepackage{mathtools}
\newcommand{\etal}{\textit{et al.}}

\hypersetup{pdfstartview={FitH}}

\title{\LARGE \bf
Deep Federated Learning for Autonomous Driving 
}
\author{Anh Nguyen$^{1,*}$, Tuong Do$^{2,*}$ Minh Tran$^{2}$, Binh X. Nguyen$^{2}$, Chien Duong$^{2}$, \\Tu Phan$^{2}$, Erman Tjiputra$^{2}$, Quang D. Tran$^{2}$
\thanks{* Equal contribution}
\thanks{$^{1}$Deparment of Computer Science, University of Liverpool, UK
        {\tt\small anh.nguyen@liverpool.ac.uk}}%
\thanks{$^{2}$AIOZ, Singapore
        {\tt\small \{tuong.khanh-long.do, minh.quang.tran, binh.xuan.nguyen, chien.duy.duong, tu.chau.phan, erman.tjiputra, quang.tran\}@aioz.io}}%
}

\begin{document}
% Macros

\newtheorem{problem}{Problem}
\newtheorem{lemma}{Lemma}
\newtheorem{theorem}[lemma]{Theorem}
\newtheorem{claim}{Claim}
\newtheorem{corollary}[lemma]{Corollary}
\newtheorem{definition}[lemma]{Definition}
\newtheorem{proposition}[lemma]{Proposition}
\newtheorem{remark}[lemma]{Remark}
\newenvironment{LabeledProof}[1]{\noindent{\it Proof of #1: }}{\qed}

\def\beq#1\eeq{\begin{equation}#1\end{equation}}
\def\bea#1\eea{\begin{align}#1\end{align}}
\def\beg#1\eeg{\begin{gather}#1\end{gather}}
\def\beqs#1\eeqs{\begin{equation*}#1\end{equation*}}
\def\beas#1\eeas{\begin{align*}#1\end{align*}}
\def\begs#1\eegs{\begin{gather*}#1\end{gather*}}

\newcommand{\poly}{\mathrm{poly}}
\newcommand{\eps}{\epsilon}
\newcommand{\e}{\epsilon}
\newcommand{\polylog}{\mathrm{polylog}}
\newcommand{\rob}[1]{\left( #1 \right)} %Round Brackets
\newcommand{\sqb}[1]{\left[ #1 \right]} %square Brackets
\newcommand{\cub}[1]{\left\{ #1 \right\} } %curly brackets
\newcommand{\rb}[1]{\left( #1 \right)} %Round
\newcommand{\abs}[1]{\left| #1 \right|} %| |
\newcommand{\zo}{\{0, 1\}}
\newcommand{\zonzo}{\zo^n \to \zo}
\newcommand{\zokzo}{\zo^k \to \zo}
\newcommand{\zot}{\{0,1,2\}}
\newcommand{\en}[1]{\marginpar{\textbf{#1}}}
\newcommand{\efn}[1]{\footnote{\textbf{#1}}}
\newcommand{\vecbm}[1]{\boldmath{#1}} %more general (handles greek letters)
\newcommand{\uvec}[1]{\hat{\vec{#1}}}
\newcommand{\thv}{\vecbm{\theta}}
\newcommand{\junk}[1]{}
\newcommand{\var}{\mathop{\mathrm{var}}}
\newcommand{\rank}{\mathop{\mathrm{rank}}}
\newcommand{\diag}{\mathop{\mathrm{diag}}}
\newcommand{\tr}{\mathop{\mathrm{tr}}}
\newcommand{\acos}{\mathop{\mathrm{acos}}}
\newcommand{\atantwo}{\mathop{\mathrm{atan2}}}
\newcommand{\SVD}{\mathop{\mathrm{SVD}}}
\newcommand{\quadf}{\mathop{\mathrm{q}}}
\newcommand{\linterp}{\mathop{\mathrm{l}}}
\newcommand{\sgn}{\mathop{\mathrm{sign}}}
\newcommand{\sym}{\mathop{\mathrm{sym}}}
\newcommand{\avg}{\mathop{\mathrm{avg}}}
\newcommand{\mean}{\mathop{\mathrm{mean}}}
\newcommand{\erf}{\mathop{\mathrm{erf}}}
\newcommand{\grad}{\nabla}
\newcommand{\R}{\mathbb{R}}
\newcommand{\defeq}{\triangleq}
\newcommand{\dims}[2]{[#1\!\times\!#2]}
\newcommand{\sdims}[2]{\mathsmaller{#1\!\times\!#2}}
\newcommand{\udims}[3]{#1}
\newcommand{\udimst}[4]{#1}
\newcommand{\com}[1]{\rhd\text{\emph{#1}}}
\newcommand{\ind}{\hspace{1em}}
\newcommand{\argmin}[1]{\underset{#1}{\operatorname{argmin}}}
\newcommand{\floor}[1]{\left\lfloor{#1}\right\rfloor}
\newcommand{\step}[1]{\vspace{0.5em}\noindent{#1}}
\newcommand{\quat}[1]{\ensuremath{\mathring{\mathbf{#1}}}}
\newcommand{\norm}[1]{\left\lVert#1\right\rVert}
\newcommand{\ignore}[1]{}
\newcommand{\specialcell}[2][c]{\begin{tabular}[#1]{@{}c@{}}#2\end{tabular}}
\newcommand*\Let[2]{\State #1 $\gets$ #2}
\newcommand{\algorithmicbreak}{\textbf{break}}
\newcommand{\Break}{\State \algorithmicbreak}
\newcommand{\ra}[1]{\renewcommand{\arraystretch}{#1}}

\renewcommand{\vec}[1]{\mathbf{#1}} %looks better

\algdef{S}[FOR]{ForEach}[1]{\algorithmicforeach\ #1\ \algorithmicdo}
\algnewcommand\algorithmicforeach{\textbf{for each}}
\algrenewcommand\algorithmicrequire{\textbf{Require:}}
\algrenewcommand\algorithmicensure{\textbf{Ensure:}}
\algnewcommand\algorithmicinput{\textbf{Input:}}
\algnewcommand\INPUT{\item[\algorithmicinput]}
\algnewcommand\algorithmicoutput{\textbf{Output:}}
\algnewcommand\OUTPUT{\item[\algorithmicoutput]}

\maketitle
\thispagestyle{empty}
\pagestyle{empty}

\begin{abstract}
Autonomous driving is an active research topic in both academia and industry. However, most of the existing solutions focus on improving the accuracy by training learnable models with centralized large-scale data. Therefore, these methods do not take into account the user's privacy. In this paper, we present a new approach to learn autonomous driving policy while respecting privacy concerns. We propose a peer-to-peer Deep Federated Learning (DFL) approach to train deep architectures in a fully decentralized manner and remove the need for central orchestration. We design a new Federated Autonomous Driving network (FADNet) that can improve the model stability, ensure convergence, and handle imbalanced data distribution problems while is being trained with federated learning methods. Intensively experimental results on three datasets show that our approach with FADNet and DFL achieves superior accuracy compared with other recent methods. Furthermore, our approach can maintain privacy by not collecting user data to a central server. Our source code can be found at: \url{https://github.com/aioz-ai/FADNet}
\end{abstract}

\section{Introduction}
Autonomous driving is an emerging field that potentially transforms the way humans travel. Most recent approaches for autonomous driving are based on machine learning, especially deep learning techniques that require large-scale training data. In particular, many works have investigated the ability to directly derive end-to-end driving policies from sensory data~\cite{pfeiffer2017perception,eraqi2017end}. The outcomes of these methods have been applied to different applications such as lane following~\cite{Gurghian16}, autonomous navigation in complex environments~\cite{hu2021sim,nguyen2020autonomous}, autonomous driving in man-made roads~\cite{Zhang17}, Unmanned Aerial Vehicles (UAV) navigation~\cite{Srivatsan19}, and robust control~\cite{chi2020collaborative,cursi2020hybrid} . 
However, most of these methods focus on improving the accuracy of the system by using pre-collected datasets rather than considering the privacy of the user data.
%Apart from supervised learning approach, reinforcement learning has been widely used to learn navigation policies from robot experiences~\cite{duan2016benchmarking, tai2017virtual, schulman2015trust,lillicrap2015continuous,wortsman2018_RL}. 

% There have been a wide range of deep learning approaches that study this problem \cite{eraqi2017end, du2019self, liang2019federated, gidado2020survey, du2020federated, pokhrel2020federated}.

While collecting data in centralized local servers would help to develop more accurate autonomous driving solutions, it strongly violates user privacy since personal data are shared with third parties. A promising solution for this problem is Federated Learning (FL). Federated Learning “\textit{involves training statistical models over remote devices or siloed data centers, such as mobile phones or hospitals, while keeping data localized}”~\cite{li2020federated}.  In practice, FL opens a new research direction where we can utilize the effectiveness of deep learning methods while maintaining the user's privacy. However, training large-scale deep networks in a decentralized way is not a trivial task~\cite{wang2018cooperative}. Furthermore, by its decentralized nature, FL comes with many challenges such as model convergence, communication congestion, or imbalance of data distributions in different silos~\cite{kairouz2019advances}.

\begin{figure}[!t] 
    \centering
    \subfigure[]{\includegraphics[width=0.2\linewidth, height=0.5\linewidth]{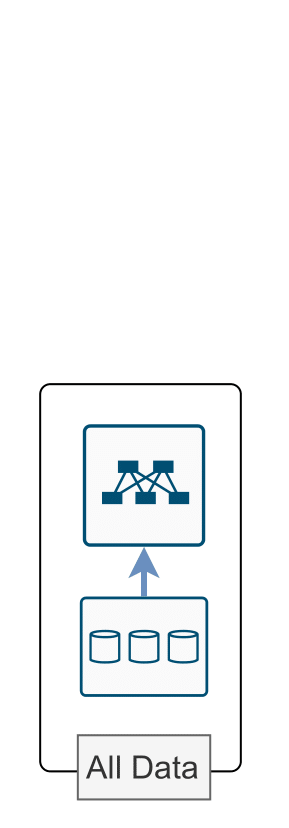}}
    \subfigure[]{\includegraphics[width=0.6\linewidth, height=0.5\linewidth]{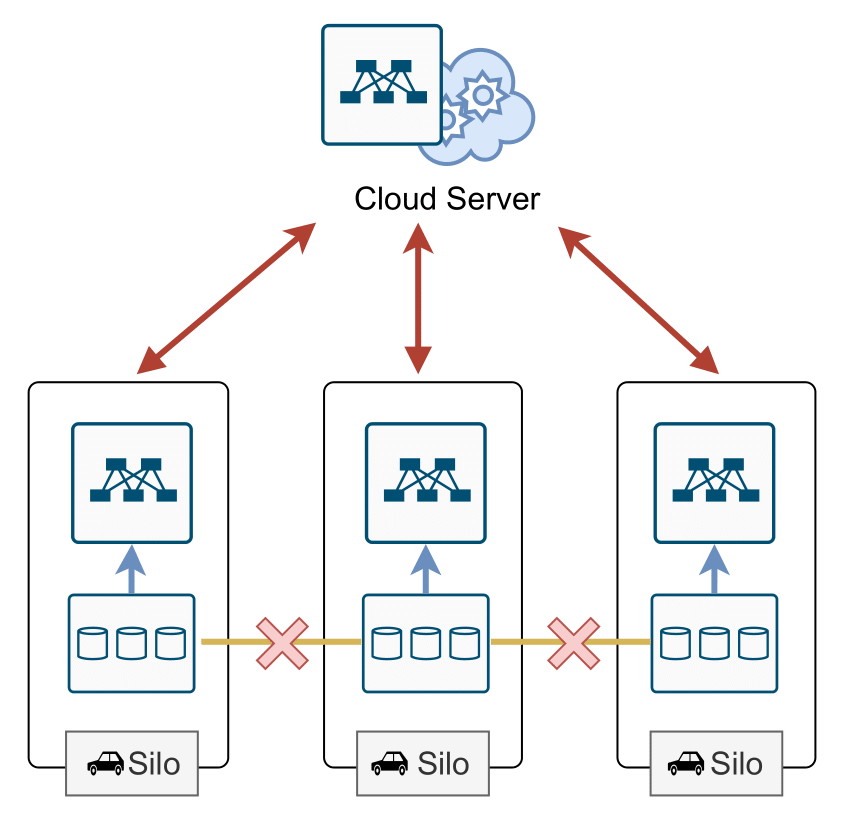}}
    \subfigure[]{\includegraphics[width=0.62\linewidth, keepaspectratio=True]{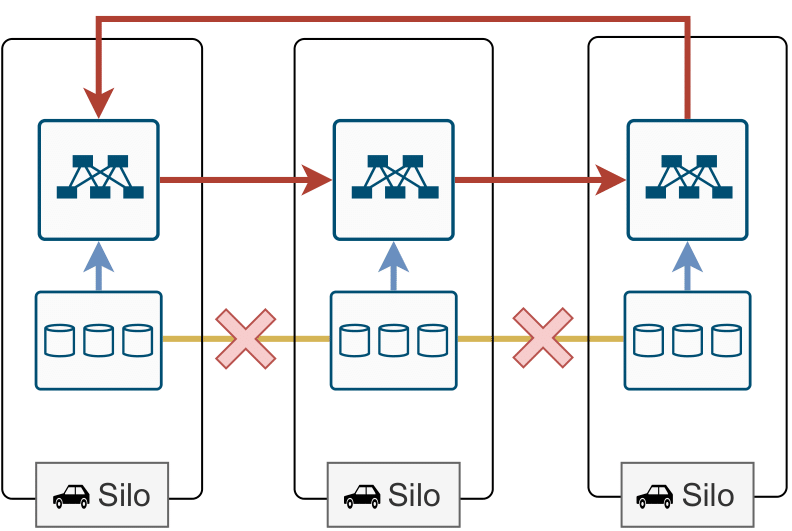}}
    %\vspace{3ex}
    \caption{An overview of different learning methods for autonomous driving. (a) Centralized Local Learning, (b) Server-based Federated Learning, and (c) our peer-to-peer Deep Federated Learning. Red arrows denote the aggregation process between silos. Yellow lines with a red cross indicate the non-sharing data between silos.}
    \label{Fig:intro} 
\end{figure}
 
In this paper, our goal is to develop an end-to-end driving policy from sensory data while maintaining the user's privacy by utilizing FL. We address the key challenges in FL to make sure our deep network can achieve competitive performance when being trained in a fully decentralized manner. Fig.\ref{Fig:intro} shows an overview of different learning approaches for autonomous driving. In \textit{Centralized Local Learning} (CLL)~\cite{bojarski2016end2end_car,loquercio2018dronet}, the data are collected and trained in one local machine. Hence, the CLL approach does not take into account the user's privacy. The \textit{Server-based Federated Learning} (SFL) strategy~\cite{sattler2019robust} requires a central server to orchestrate the training process and receive the contributions of all clients. The main limitation of SFL is communication congestion when the number of clients is large. Therefore, we follow the peer-to-peer federated learning~\cite{marfoq2020throughput} to set up the training. Our peer-to-peer \textit{Deep Federated Learning} (DFL) is fully decentralized and can reduce communication congestion during training. We also propose a new Federated Autonomous Driving network (FADNet) to address the problem of model convergence and imbalanced data distribution. By training our FADNet using DFL, our approach outperforms recent state-of-the-art methods by a fair margin while maintaining user data privacy.

Our contributions can be summarized as follows:
\begin{itemize}
    \item We propose a fully decentralized, peer-to-peer Deep Federated Learning framework for training autonomous driving solutions.
    \item We introduce a Federated Autonomous Driving network that is well suitable for federated training.
    \item We introduce two new datasets and conduct intensive experiments to validate our results.
\end{itemize}
% \textbullet{~We propose the training and inference support with Distributed Federated Learning (DFL) in the task of predicting steering angles. The performance is comparable with local state-of-the-art training methods.}

% \textbullet{~We simulate a read-world scenarios by collecting data from distinct cities and weathers. Thereby, we extensively carry out the experiments to prove the effectiveness of our methods.}

% \textbullet{~We deploy our methods in a real-world robot. Experiments show that ...}

% The remainder of this paper is organized as follows:

\section{Related works}

\textbf{Deep Learning.} Deep learning is a popular approach to learn an end-to-end driving policy from sensory data~\cite{xu2017end, richter2017safe,amini2018learning,nguyen2021autonomous}. Bojarski \etal~\cite{bojarski2016end2end_car} introduced a deep network for autonomous driving with inputs from 2D images. The authors in~\cite{smolyanskiy2017toward} developed a deep navigation network for UAVs using images from three cameras. In~\cite{loquercio2018dronet}, the authors used a deep network to learn the navigation policy and predict the collision probability. In~\cite{amini2018variational}, a deep network was combined with Variational Autoencoder to estimate the steering angle. The work of~\cite{alexander_2019_variational_end2end,du2019self} built the navigation map from visual inputs to learn the control policy. More recently, deep learning is widely applied to solve various problems in autonomous driving such as 3D object detection, visual question answering~\cite{nguyen2021coarse}, and obstacle avoidance~\cite{choi2021shared, badki2021binary,chen2021multisiam,nguyen2019scene,ren2021safetyaware}. The authors in~\cite{liu2021ground} investigated how the ground plane contributes to 3D detection in driving scenarios. In~\cite{prakash2021multi}, the authors proposed fusion transformer for autonomous driving. Reinforcement learning and adversarial learning~\cite{tran2022light} have also been widely used to learn driving policies~\cite{wortsman2018_RL, piotr_2017_navigate_in_complex_environments,delbrouck2018_object_navigation}.

%The authors in~\cite{zhu2017target} proposed a RL solution for the target-driven navigation problem. The authors in~\cite{delbrouck2018_object_navigation} utilized semantic information and spatial relationships for target objects navigation. In~\cite{muller2018teaching}, an end-to-end regression system was introduced for UAV racing. 

\textbf{Federated Learning.} Federated learning has received much attention from the research community recently. Numerous FL approaches have been introduced for different applications such as finance~\cite{shingi2020federated}, healthcare~\cite{xu2021federated}, and medical image~\cite{courtiol2019deep}. To train an FL method, the cross-silo approach has become popular since it enables the training process to use computing resources more effectively~\cite{marfoq2020throughput}.
%FedAvg \cite{mcmahan2017communication}, SCAFFOLD \cite{karimireddy2020scaffold}, and FedProx \cite{li2018federated} proposed orchestrator-centered communications. 
The authors in~\cite{li2021decentralized} proposed a framework called decentralized federated learning via mutual knowledge transfer. Liu \etal~\cite{liu2019federated} introduced a knowledge fusion algorithm for FL in cloud robotics. In autonomous driving, Zhang \etal~\cite{zhang2021real} proposed a real-time end-to-end FL approach that includes a unique asynchronous model aggregation mechanism. The authors in~\cite{doomra2020turn} used FL to predict the turning signal. More recently, the authors in~\cite{barbieri2021decentralized,khan2021dispersed} used FL for 6G-enabled autonomous cars. Peng \etal~\cite{peng2021bflp} introduced an adaptive FL framework for autonomous vehicles. In~\cite{zhang2021distributed}, the authors addressed the problem of distributed dynamic map fusion with FL for intelligent networked vehicles.

Unlike other approaches that focus on improving the robustness of autonomous driving solutions while ignoring the user's privacy. In this work, we propose to address the autonomous driving problem using federated learning. Our method contains a decentralized federated learning algorithm co-operated with our specialized network design, which can improve accuracy and reduce communication congestion during the distributed training process. 

\section{Methodology}
\subsection{Problem Formulation}
We consider a federated network with $N$ siloed data centers (e.g., autonomous cars) $\mathcal{D}_{i}$, with $i \in [1,N]$. Our goal is to collaboratively train a global driving policy $\theta$ by aggregating all local learnable weights $\theta_i$ of each silo. Note that, unlike the popular centralized local training setup~\cite{loquercio2018dronet}, in FL training, each silo does not share its local data, but periodically transmits model updates to other silos~\cite{karimireddy2020scaffold,wang2018cooperative}.
% \begin{equation}
% \label{eq:general_opt}
%     \theta = \frac{1}{N}\sum^N_{i=1} \mathcal{L}_i(\xi_i, \theta_i)
% \end{equation}

In practice, each silo has the training loss $\mathcal{L}_i(\xi_i, \theta_i)$. $\xi_i$ is the ground-truth in each silo $i$.  $\mathcal{L}_i(\xi_i, \theta_i)$ is calculated as the regression loss. This regression loss is modeled by a deep network that takes RGB images as inputs and predicts the associated steering angles~\cite{loquercio2018dronet}.

% Each one contains experiments recorded from navigating agents. More concretely, observations $\{o_0^{(i)}, o_1^{(i)} ...o_T^{(i)}\}$ and ground truths $\{a_0^{(i)}, a_1^{(i)}, ..., a_T^{(i)}\}$ are provided, where $T$ is the terminated timestamp.

% At timestamp $t$, let $o_t$ is the observation of the agent. The problem is to learn a navigating policy $\pi_{\theta}$ which predicts the steering angle $a_t$.
% \begin{equation}
%     a_t = \pi_{\theta} (o_t)
% \end{equation}

% This prediction are supervised by the corresponding ground truth, $a^{gt}_t$. The supervised loss function $\mathcal{L}$ is the regression loss (mean square errors) $L_{MSE}$ for steering actions.
% \begin{equation}
%     \mathcal{L} = L_{MSE}(a_t, a^{gt}_t)
% \end{equation}

\subsection{Deep Federated Learning for Autonomous Driving}

\begin{figure*}[!h]
  \centering
%   \subfigure[Typical local training. ]{\includegraphics[width=0.29\linewidth]{images/Local.png}}
  \subfigure[]{\includegraphics[width=0.25\linewidth]{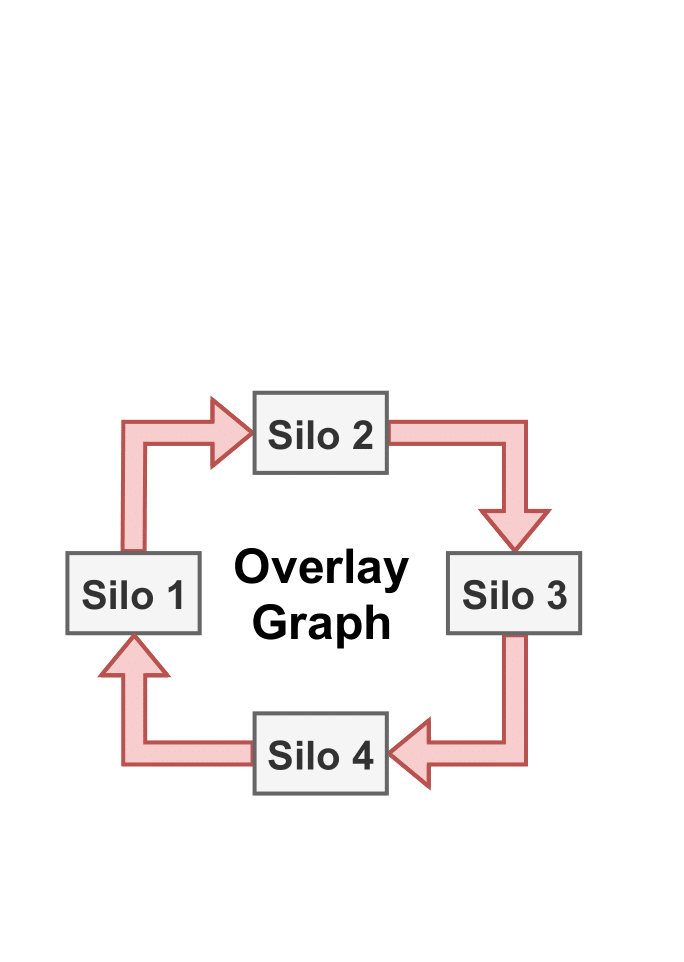}}
  \subfigure[]{\includegraphics[width=0.7\linewidth]{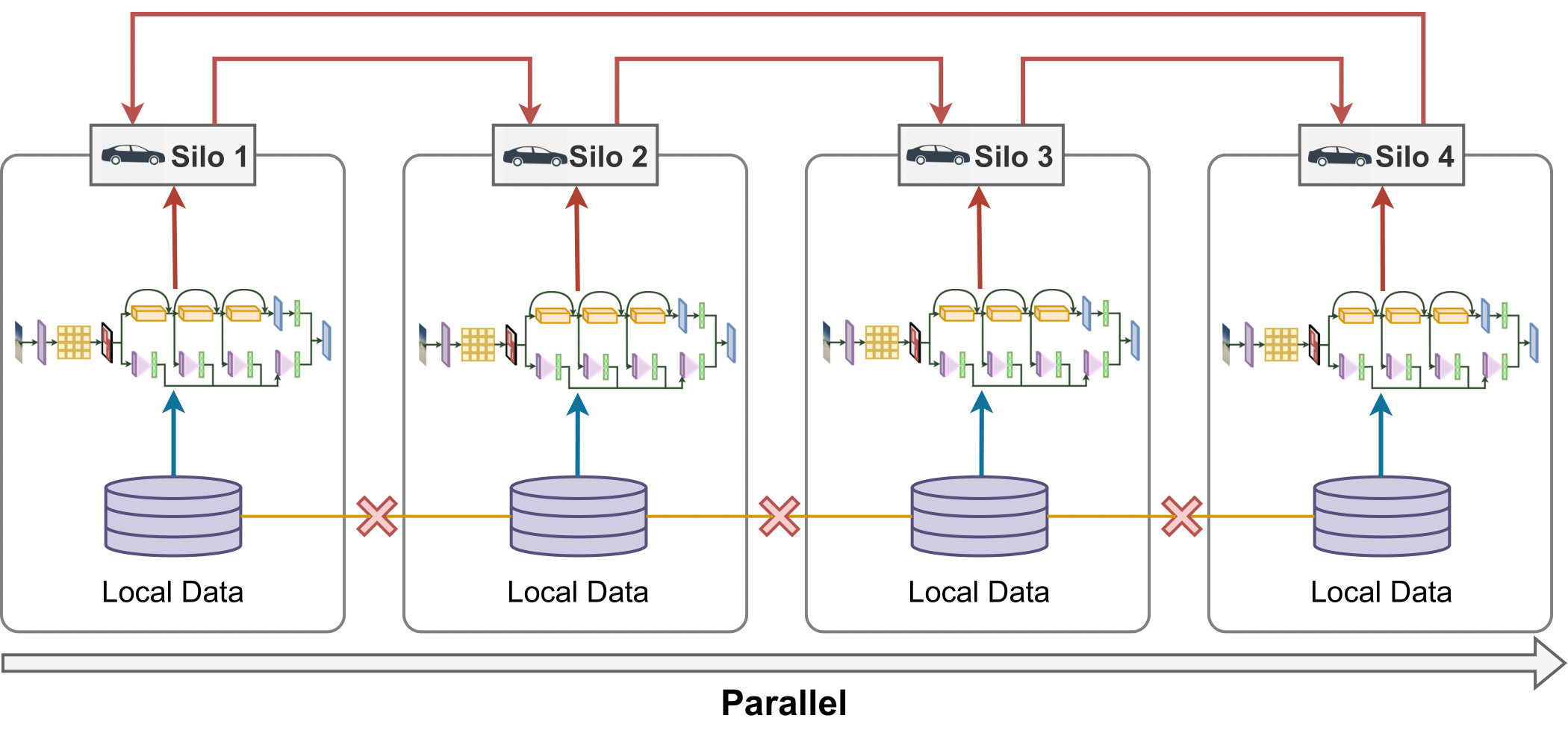}}
% %   \subfigure[Supporting Topology for Decentralized Federated Learning]{\includegraphics[width=0.24\linewidth]{images/Topo.png}}
    % \vspace{0.1ex}
    % \includegraphics[width=1.0\linewidth]{images/DFL_v4.png}
\vspace{-1ex}
 \caption{An overview of our peer-to-peer Deep Federated Learning method. (a) A simplified version of an overlay graph. (b) The training methodology in the overlay graph. Note that blue arrows denote the local training process in each silo; red arrows denote the aggregation process between silos controlled by the overlay graph; yellow lines with a red cross indicate the non-sharing data between silos; the arrow indicates that the process is parallel.}
 \label{fig:learning_methods}
  \vspace{-1ex}
\end{figure*}

A popular training method in FL is to set up a central server that orchestrates the training process and receives the contributions of
all clients (Server-based Federated Learning - SFL)~\cite{sattler2019robust}. The limitation of SFL is the server potentially represents a single point of failure in the system. We also may have communication congestion between the server and clients when the number of clients is massive~\cite{lian2017can}. Therefore, in this work, we utilize the peer-to-peer FL~\cite{marfoq2020throughput} to set up the training scenario. In peer-to-peer FL, there is no centralized orchestration, and the communication is via peer-to-peer topology. However, the main challenge of peer-to-peer FL is to assure model convergence and maintain accuracy in a fully decentralized training setting.

% DFL training algorithm aims to deal with the silo bias during training and the server congestion problem in CFL by leveraging the overlay graph design. Specifically, silos connect to their neighbors to transfer their learned weights frequently. Thus, the bias in some specific silos is reduced.
% Besides, overlay enables each silo to work as an orchestrator. Thus, no server is required, and then training speed is increase. Inspired by~\cite{marfoq2020throughput}

Fig.~\ref{fig:learning_methods} illustrates our Deep Federated Learning (DFL) method. Our DFL follows the peer-to-peer FL setup with the goal to integrate a deep architecture into a fully decentralized setting that ensures convergence while achieving competitive results compared to the traditional Centralized Local Learning~\cite{loquercio2018dronet} or SFL~\cite{sattler2019robust} approach. In practice, we can consider a \textit{silo} as an autonomous car. Each silo maintains a local learnable model and does not share its data with other silos. We represent the silos as vertices of a communication graph and the FL is performed on an \textit{overlay}, which is a sub-graph of this communication graph.

\subsubsection{Designing the Overlay}
Let $\mathcal{G}_c = (\mathcal{V}, \mathcal{E}_c)$ is the connectivity graph that captures the possible direct communications among $N$ silos. $\mathcal{V}$ is the set of vertices (silos), while $\mathcal{E}_c$ is the set of communication links between vertices. $\mathcal{N}_i^{+}$ and $\mathcal{N}_i^{-}$ are in-neighbors and out-neighbors of a silo $i$, respectively. As in~\cite{marfoq2020throughput}, we note that it is unnecessary to use all the connections of the connectivity graph for FL. Indeed, a sub-graph called an overlay, $\mathcal{G}_o = (\mathcal{V}, \mathcal{E}_o)$ can be generated from $\mathcal{G}_c$. In our work, $\mathcal{G}_o$ is the result of Christofides’ Algorithm~\cite{monnot2003approximation}, which yields a strong spanning sub-graph of $\mathcal{G}_c$ with minimal cycle time. One cycle time or time per communication round, in general, is the time that a vertex waits for messages from the other vertices to do a computational update. The computational update will be described in the training algorithm in Section~\ref{training_procedure}. 

In practice, one block cycle time of an overlay $\mathcal{G}_o$ depends on the delay of each link $(i, j)$, denoted as $d_o(i, j)$, which is the time interval between the beginning of a local computation at node $i$, and the receiving of $i$'s messages by $j$. Furthermore, without concerns about access links delays between vertices, our graph is treated as an edge-capacitated network with:
\begin{equation}
\label{delay_d}
    d_o(i,j) = s \times T_c(i) + l(i,j) + \frac{M}{B(i,j)}
\end{equation}
where $T_c(i)$ is the time to compute one local update of the model; $s$ is the number of local computational steps; $l(i,j)$  is the link latency; $M$ is the model size; $B(i,j)$ is available bandwidth of the path $(i,j)$. As in~\cite{marfoq2020throughput},  we set $s=1$.

\begin{figure*}[!ht]
   \centering
   \includegraphics[width=1.0\linewidth]{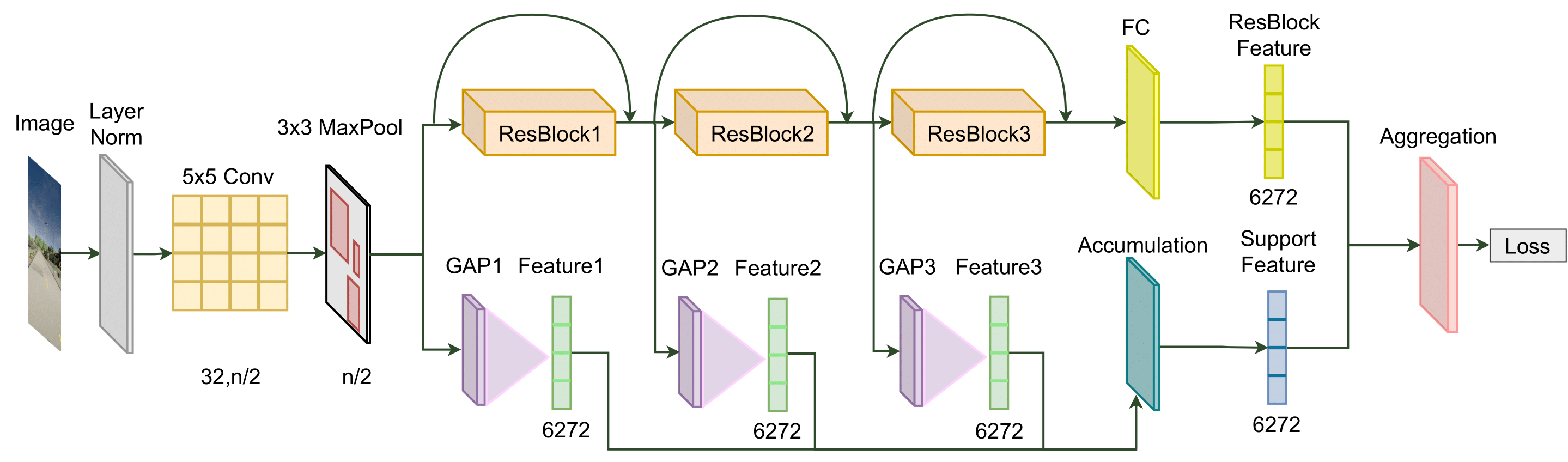}
  % \vspace{3ex}
 \caption{The architecture of our Federated Autonomous Driving Net (FADNet).  
 }
 \label{fig:network}
\vspace{-1ex}
\end{figure*}

\subsubsection{Training Algorithm}
\label{training_procedure}
At each silo $i$, the optimization problem to be solved is:
\begin{equation}
\label{eq:theta_i_computation}
    \theta_i^{*} = \underset{\theta_i}{\arg\min} \underset{\xi \sim \mathcal{D}_i}{\mathbb{E}}[\mathcal{L}(\xi_i, \theta_i)]
\end{equation}

We apply the distributed federated learning algorithm, DPASGD~\cite{wang2018cooperative}, to solve the optimizations of all the silos. In fact, after waiting one cycle time, each silo $i$ will receive parameters $\theta_j$ from its in-neighbor $\mathcal{N}_i^{+}$ and accumulate these parameters multiplied with a non-negative coefficient from the consensus matrix $\mathbf{A}$. It then performs $s$ mini-batch gradient updates before sending $\theta_i$ to its out-neighbors $\mathcal{N}_i^{-}$, and the algorithm keeps repeating. Formally, at each iteration $k$, the updates are described as: 

% \begin{equation}
% \theta_{i}\left(k + 1\right) = 
% \begin{cases}
%     \sum_{j \in \mathcal{N}_i^{+} \cup{i}}\textbf{A}_{i,j}{\theta}_{j}\left(k\right), & \textit{if k} \equiv 0 \pmod{s + 1},\\
%     {\theta}_{i}\left(k\right)-\alpha_{k}\frac{1}{m}\sum^m_{h=1}\nabla \mathcal{L}\left({\theta}_{i}\left(k\right),\xi_i^{\left(h\right)}\left(k\right)\right), & \text{otherwise.}
% \end{cases}
% \label{eq:ori_DFL}
% \end{equation}

 \begin{equation}
\begin{aligned}
 &\theta_{i}\left(k + 1\right) \\
 & =\begin{cases}
    \sum_{j \in \mathcal{N}_i^{+} \cup{i}}\textbf{A}_{i,j}{\theta}_{j}\left(k\right), \textit{ \quad \quad \quad if k} \equiv 0 \pmod{s + 1},\\
    {\theta}_{i}\left(k\right)-\alpha_{k}\frac{1}{m}\sum^m_{h=1}\nabla \mathcal{L}\left({\theta}_{i}\left(k\right),\xi_i^{\left(h\right)}\left(k\right)\right),  \text{otherwise.}
\end{cases}
\end{aligned}
\label{eq:ori_DFL}
\end{equation}
where $m$ is the mini-batch size and $\alpha_k > 0$ is a potentially varying learning rate.

\subsubsection{Federated Averaging}
Following~\cite{mcmahan2017communication}, to compute the prediction of models in all silos, we compute the average model $\theta$ using weight aggregation from all the local model $\theta_i$. The federated averaging process is conducted as follow:
\begin{equation}
\theta = \frac{1}{\sum^N_{i=0}{\lambda_i}} \sum^N_{i=0}\lambda_{{i}} \theta_{{i}}
\label{eq:aggr_model}
\end{equation}
where $N$ is the number of silos; $\lambda_i = \{0,1\}$. Note that $\lambda_i = 1$ indicates that silo $i$ joins the inference process and $\lambda_i = 0$ if not. The aggregated weight $\theta$ is then used for evaluation on the testing set $\mathcal{D}_{test}$.

\subsection{Network Architecture}

According to~\cite{sattler2019robust,wang2018cooperative}, one of the main challenges when training a deep network in FL is the imbalanced and non-IID (identically and independently distributed) problem in data partitioning across silos. To overcome this problem, the learning architecture should have an appropriate design to balance the trade-off between convergence ability and accuracy performance. In practice, the deep architecture has to deal with the high variance between silo weights when the accumulation process for all silos is conducted. To this end, we design a new Federated Autonomous Driving Network, which is based on ResNet8~\cite{he2016deep}, as shown in Fig.~\ref{fig:network}. 

% Besides, the network should have enough complexity to achieve optimized predicted results.

In particular, our proposed FADNet first comprises an input layer normalization to improve the stability of the abstract layer. This layer aims to handle different distributions of input images in each silo. Then, a convolution layer following by a max-pooling layer is added to encode the input. To handle the vanishing gradient problem, three residual blocks~\cite{he2016deep} are appended with a following FC layer to extract ResBlock features. However, using residual blocks increases the variance of silo weights during the aggregation process and affects the convergence ability of the model. To address this problem, we add a Global Average Pooling layer (GAP)~\cite{lin2013network} associated with each residual block. GAP is a non-weight pooling layer which sums out the spatial information from each residual block. Thus, it is not affected by the weighted variance problem. The output of each GAP layer is passed through an Accumulation layer to accrue the Support feature. 
The ResBlock feature and the Support feature from GAP layers are fed into the Aggregation layer to calculate the model loss in each silo.

In our design, the Accumulation and Aggregation layers aim to reduce the variance of the global model since we need to combine multiple model weights produced by different silos. In particular, the Accumulation layer is a variant of the fully connected (FC) layer. Instead of weighting the contribution of input nodes as in FC, the Accumulation layer weights the contribution of multiple features from input layers. The Accumulation layer has a learnable weight matrix $w \in \mathbb{R}^\text{n}$. Its number of nodes is equal to the \text{n} number of input layers. Note that the support feature from the Accumulation layer has the same size as the input. Let $F = \{f_\text{1}, f_\text{2}, ..., f_\text{n}\}, \forall f_\text{h} \in \mathbb{R}^\text{d}$ be the collection of $\text{n}$ number of the features extracted from $\text{n}$ input GAP layers; $\text{d}$ is the unified dimension. The Accumulation outputs a feature $f_\text{c} \in \mathbb{R}^\text{d}$ in each silo $i$, and is computed as:
\begin{equation}
f_\text{c} = Accumulation(F)_i = \sum^{\text{n}}_{\text{h}=1}(w_\text{h}f_\text{h})_i
\label{eq:accu}
\end{equation}

The Aggregation layer is a fusion between the ResBlock feature extracted from the backbone and the support feature from the Accumulation layer. For simplicity, we use the Hadamard product to compute the aggregated feature. This feature is then averaged to predict the steering angle. Let $f_\text{s} \in \mathbb{R}^\text{d}$ be the ResBlock features extracted from the backbone. The output driving policy $\theta_i$ of silo $i$ can be calculated as:
\begin{equation}
\theta_i = Aggregation(f_\text{s}, f_\text{c})_i = \bar{(f_\text{s} \odot f_\text{c})}_i
\label{eq:aggr}
\end{equation}
where $\odot$ denotes Hadamard product; $\bar{(*)}$ denotes the mean. As in~\cite{loquercio2018dronet}, we set $\text{d} = 6,272$.

\section{Results}

\subsection{Experimental Setup}
\textbf{Udacity.} We use the popular Udacity dataset~\cite{udacity2016data} to evaluate our results. We only use front-forwarded images in this dataset in our experiment. As in~\cite{loquercio2018dronet}, we use $5$ sequences for training and $1$ for testing. The training sequences are assigned randomly to different silos depending on the federated topology (i.e., Gaia~\cite{knight2011internetzoo} or NWS~\cite{awscloud}).

\begin{figure}[!h]
   \centering
\resizebox{\linewidth}{!}{
\setlength{\tabcolsep}{1pt}
\begin{tabular}{ccc}

\shortstack{\includegraphics[width=0.33\linewidth]{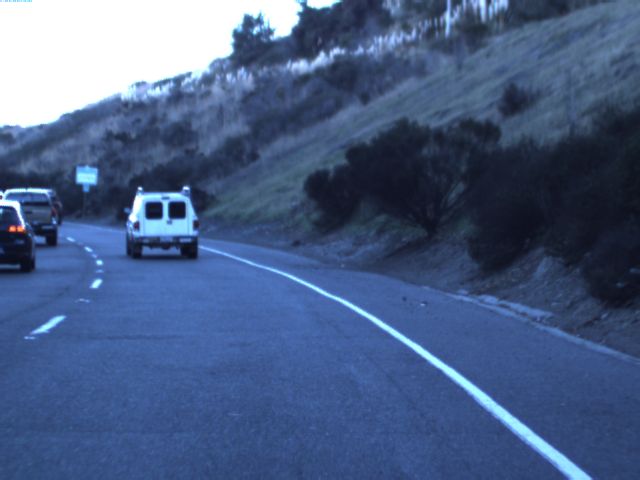}}&
% \shortstack{\includegraphics[width=0.3\linewidth]{images/DatasetSamples/Udacity/2.jpg}}&
\shortstack{\includegraphics[width=0.33\linewidth]{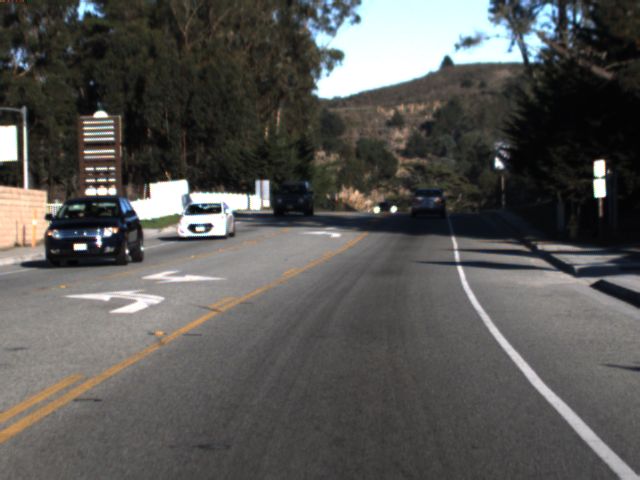}}&
% \shortstack{\includegraphics[width=0.3\linewidth]{images/DatasetSamples/Udacity/4.jpg}}&
% \shortstack{\includegraphics[width=0.3\linewidth]{images/DatasetSamples/Udacity/5.jpg}}&
\shortstack{\includegraphics[width=0.33\linewidth]{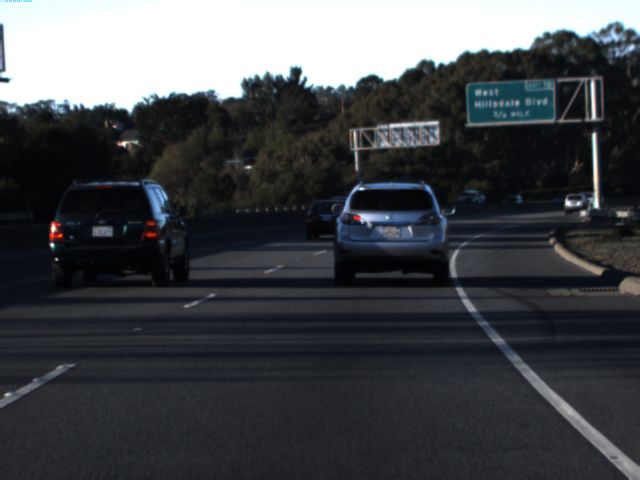}}\\ [0pt]
% \hline\\
% \shortstack{\includegraphics[width=0.3\linewidth]{images/DatasetSamples/Indoor/1.jpg}}&
\shortstack{\includegraphics[width=0.33\linewidth]{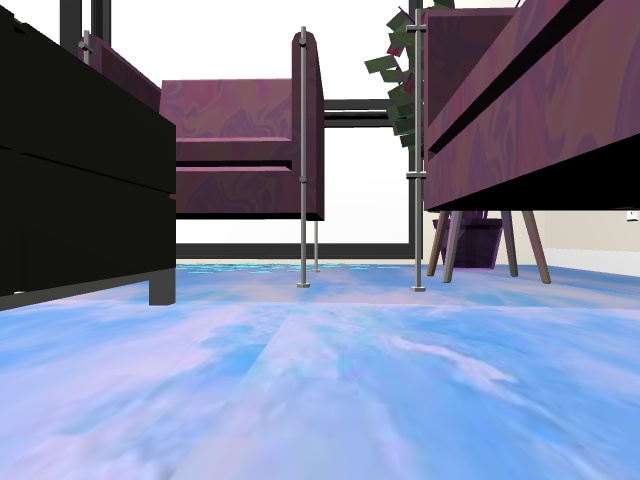}}&
% \shortstack{\includegraphics[width=0.3\linewidth]{images/DatasetSamples/Indoor/3.png}}&
% \shortstack{\includegraphics[width=0.3\linewidth]{images/DatasetSamples/Indoor/4.jpg}}&
\shortstack{\includegraphics[width=0.33\linewidth]{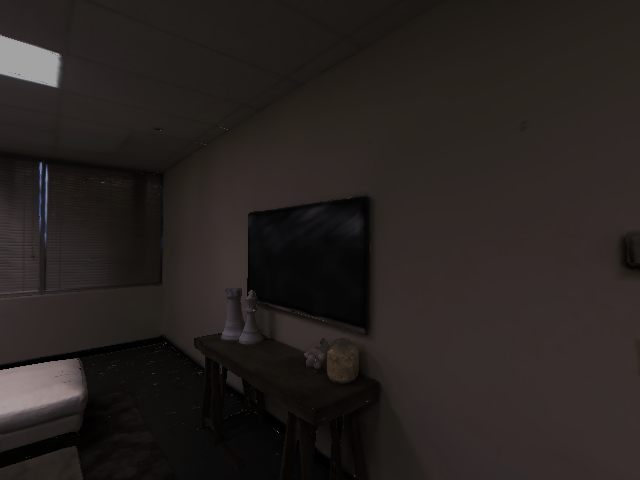}}&
\shortstack{\includegraphics[width=0.33\linewidth]{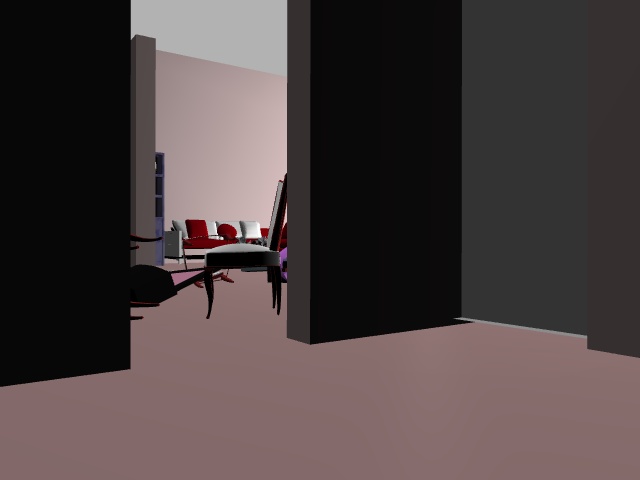}}\\ [0pt]
% \hline\\
\shortstack{\includegraphics[width=0.33\linewidth]{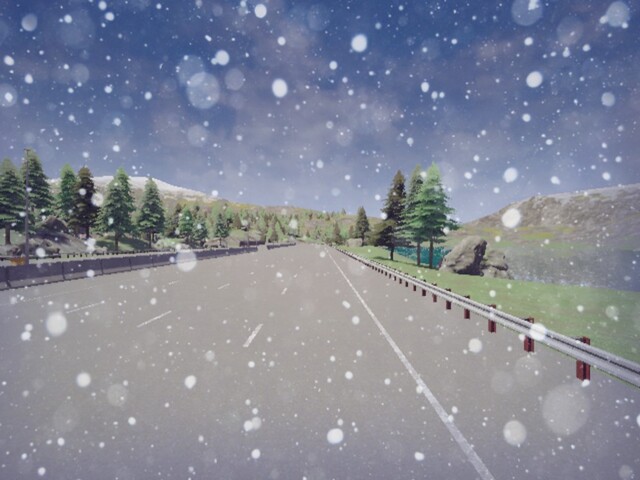}}&
\shortstack{\includegraphics[width=0.33\linewidth]{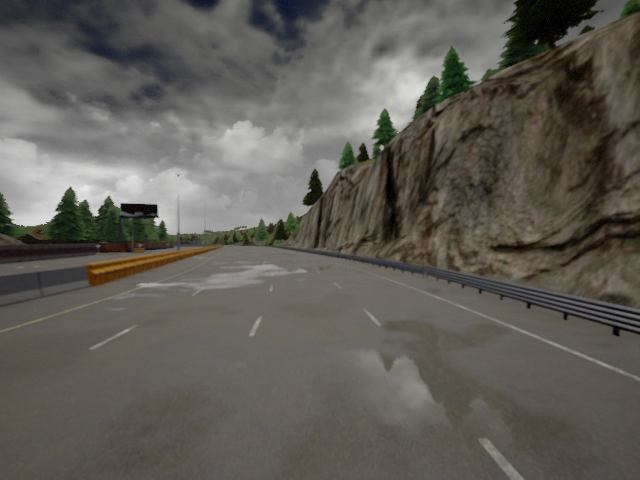}}&
% \shortstack{\includegraphics[width=0.3\linewidth]{images/DatasetSamples/Outdoor/3.jpg}}&
% \shortstack{\includegraphics[width=0.3\linewidth]{images/DatasetSamples/Outdoor/4.jpg}}&
% \shortstack{\includegraphics[width=0.3\linewidth]{images/DatasetSamples/Outdoor/5.jpg}}&
\shortstack{\includegraphics[width=0.33\linewidth]{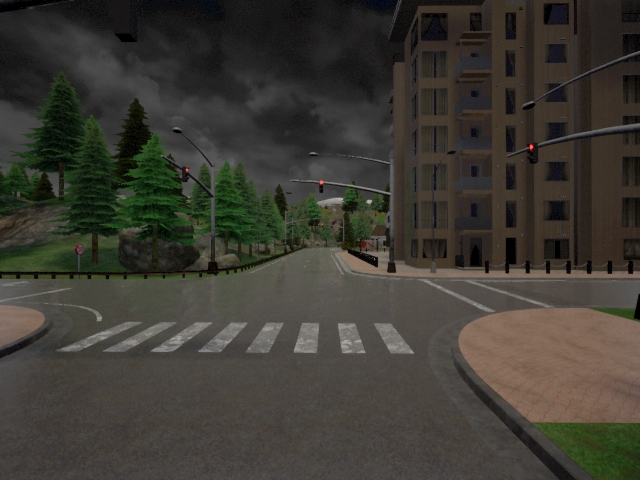}}\\ [0pt]
\end{tabular}
}
%\vspace{3ex}
    \caption{Visualization of sample images in three datasets: Udacity (first row), Gazebo (second row), and Carla (third row).}
    \label{fig:three_datasets_samples}
\end{figure}

\textbf{Carla.} Since the Udacity dataset is collected in the real-world environment, changing the weather or lighting conditions is not easy. To this end, we collect more simulated data in the Carla simulator. We have applied different lighting (morning, noon, night, sunrise, sunset) and weather conditions (cloudy, rain, heavy rain, wet streets, windy, snowy) when collecting the data. We have generated $73,235$ samples distributed over $11$ sequences of scenes. 

\textbf{Gazebo.} Since both the Udacity and Carla datasets are collected in outdoor environments, we also employ Gazebo to collect data for autonomous navigation in indoor scenes. We use a simulated mobile robot and the built-in scenes introduced in~\cite{nguyen2020autonomous} to collect data. Table~\ref{tab:datasets} shows the statistics of three datasets. We use $80\%$ of the collected data in Gazebo and Carla data for training, and the rest $20\%$ for testing.

%To meet the Non-IID requirement in Gazebo and Carla datasets during the training/testing phase, the train/validate/test set are split by data of each silo. Specifically, in Gaia network, train set includes data from $1$-st silo to $9$-th silo, validate set contains data from $10$-th silo, and test set is the whole data of $11$-th silo. In NWS network, train set is combined using data from $1$-st silo to $18$-th silo, validate set have the whole data of $19$-th and $20$-th silo, and test set holds data of $21$-th and $22$-th silo.

\begin{table}[!h]
\begin{center}
%\footnotesize
\small
% \vspace{+0.15 cm}
\setlength{\tabcolsep}{0.5 em} % for the horizontal padding
{\renewcommand{\arraystretch}{1.2}% for the vertical padding
\begin{tabular}{|c|c|c|c|}
\hline
\multirow{3}{*}{\textbf{Dataset}} & \multirow{3}{*}{\textbf{\begin{tabular}[c]{@{}c@{}}Total \\ samples\end{tabular}}} & \multicolumn{2}{c|}{\textbf{\begin{tabular}[c]{@{}c@{}}Average samples  in each silo\end{tabular}}}                                                                            \\ \cline{3-4} 
                                  &                                                                                    & \multirow{2}{*}{\textit{\begin{tabular}[c]{@{}c@{}}Gaia~\cite{knight2011internetzoo}\\ (11 silos)\end{tabular}}} & \multirow{2}{*}{\textit{\begin{tabular}[c]{@{}c@{}}NWS~\cite{awscloud}\\ (22 silos)\end{tabular}}} \\
                                  & & &  \\ \hline
Udacity                           & 39,087                                                                                   & 3,553                                                                                      & 1,777                                                                                   \\ \hline
Gazebo                       & 66,806                                                                                    &  6,073                                                                                    & 3,037                                                                                   \\ \hline
Carla                          & 73,235                                                                                   & 6,658                                                                                     & 3,329                                                                                   \\ \hline
\end{tabular}
}
\end{center}
% % \vspace{-0.5cm}
\caption{The Statistic of Datasets in Our Experiments.
}
\label{tab:datasets}
\end{table}

\textbf{Network Topology.} Following~\cite{marfoq2020throughput}, we conduct experiments on two topologies: the Internet Topology Zoo~\cite{knight2011internetzoo} (Gaia), and the North America data centers~\cite{awscloud} (NWS). We use Gaia topology in our main experiment and provide the comparison of two topologies in our ablation study. 

\textbf{Training.} The model in a silo is trained with a batch size of $32$ and a learning rate of $0.001$ using Adam optimizer. We follow the training process (Section~\ref{training_procedure}) to obtain a global weight of all silos. The training process is conducted with $3000$ communication rounds and each silo has one NVIDIA 1080 11 GB GPU for training. Note that, one communication round is counted each time all silos have finished updating their model weights. The approximate training time of all silos is shown in Fig.~\ref{fig:graph}.  For the decentralized testing process, we use the Weight Aggregation method~\cite{wang2018cooperative}.

% \begin{figure}[!h]
%   \centering
%   \includegraphics[width=1.0\linewidth]{images/SteerAngDis/his_all.png}
% % \resizebox{\linewidth}{!}{
% % \setlength{\tabcolsep}{2pt}
% %\setlength{\tabrowsep}{12pt}
% % \begin{tabular}{ccc}
% % \shortstack{\includegraphics[width=0.33\linewidth]{images/SteerAngDis/steer_historgram2.png}}&
% % \shortstack{\includegraphics[width=0.33\linewidth]{images/SteerAngDis/steer_historgram8.png}}&
% % \shortstack{\includegraphics[width=0.33\linewidth]{images/SteerAngDis/steer_historgram10.png}}\\[1pt]
% % \end{tabular}
% % }
%     \caption{\textbf{To discuss later} Sample viewpoints over $3$ different silos accompanied with their steering angle distributions in our Carla dataset.}
%     \label{fig:10_silos}
% \end{figure}

\textbf{Baselines.}
We compare our results with various recent methods, including Random baseline and Constant baseline~\cite{loquercio2018dronet}, Inception-V3~\cite{szegedy2016rethinking}, MobileNet-V2~\cite{sandler2018mobilenetv2}, VGG-16~\cite{simonyan2014very}, and Dronet~\cite{loquercio2018dronet}. All these methods use the Centralized Local Learning (CLL) strategy (i.e., the data are collected and trained in one local machine.) For distributed learning, we compare our Deep Federated Learning (DFL) approach with the Server-based Federated Learning (SFL) strategy~\cite{sattler2019robust}. As the standard practice~\cite{loquercio2018dronet}, we use the root-mean-square error (RMSE) metric to evaluate the results.

\subsection{Results}

\begin{table}[h]
\begin{center}
%\footnotesize
\small
% \vspace{+0.15 cm}
\setlength{\tabcolsep}{0.22 em} % for the horizontal padding
{\renewcommand{\arraystretch}{1.2}% for the vertical padding
\begin{tabular}{|c|c|c|c|c|r|}
\hline
\multirow{2}{*}{\textbf{Architecture}} & \multirow{2}{*}{\textbf{\begin{tabular}[c]{@{}c@{}}Learning\\Method\end{tabular}}} & \multicolumn{3}{c|}{\textbf{Dataset}}                                     & \multirow{2}{*}{\textbf{\#Params}} \\ \cline{3-5}
                                       &                                     & \textit{\textbf{Udacity}} & \textit{\textbf{Gazebo}} & \textit{\textbf{Carla}} &                                     \\ \hline
Random~\cite{loquercio2018dronet}                                & -                                 & 0.301                     & 0.117                 & 0.464                 & \_                                  \\ \hline
Constant~\cite{loquercio2018dronet}                              & -                                 & 0.209                     & 0.092                 & 0.348                 & \_                                  \\ \hline
Inception~\cite{szegedy2016rethinking}                          & CLL                                 & 0.154                     & 0.085                 & 0.297                 & 21,787,617                            \\ \hline
MobileNet~\cite{sandler2018mobilenetv2}                          & CLL                                 & 0.142                     & 0.083                 & 0.286                 & 2,225,153                             \\ \hline
VGG-16~\cite{simonyan2014very}                                & CLL                                 & 0.121                     & 0.083                 & 0.316                 & 7,501,587                             \\ \hline
DroNet~\cite{loquercio2018dronet}                                & CLL                                 & 0.110                     & 0.082                 & 0.333                 & 314,657                              \\ \hline
FADNet (ours)                          & DFL                        & \textbf{0.107}            & \textbf{0.069}        & \textbf{0.203}        & 317,729                     \\ \hline
\end{tabular}
}
\end{center}
% % \vspace{-0.5cm}
\caption{Performance comparison between different methods. The Gaia network topology is used in our DFL learning method.
}
\label{tab:sota}
\end{table}
Table~\ref{tab:sota} summarises the performance of our method and recent state-of-the-art approaches. We notice that our FADNet is trained using the proposed peer-to-peer DFL using the Gaia topology with 11 silos. This table clearly shows our FDANet + DFL outperforms other methods by a fair margin. In particular, our FDANet + DFL significantly reduces the RMSE in Gazebo and Carla datasets, while slightly outperforms DroNet in the Udacity dataset. These results validate the robustness of our FADNet while is being trained in a fully decentralized setting. Table~\ref{tab:sota} also shows that with a proper deep architecture such as our FADNet, we can achieve state-of-the-art accuracy when training the deep model in FL. Fig.~\ref{fig:act_map} illustrates the spatial support regions when our FADNet making the prediction. Particularly, we can see that FADNet focuses on the ``line-like" patterns in the input frame, which guides the driving direction.

% \begin{figure}[!h]
%   \centering
%   \subfigure[]{\includegraphics[width=0.30\linewidth]{images/Vis/Uda_ActMap.jpg}}
%   \subfigure[]{\includegraphics[width=0.30\linewidth]{images/Vis/NVO_ActMap.jpg}}
%   \subfigure[]{\includegraphics[width=0.30\linewidth]{images/Vis/NVI_ActMap.jpg}}
%     % \vspace{0.1ex}
%  \caption{\textbf{Should add original image on top?} Spatial support regions for steering regression in three different dataset. (a) Real image Udacity dataset; (b) Non I.I.D image data in NVO dataset; (c) Indoor image in NVI dataset.  In most cases, ee can observe that the model focus on “line-like” patterns to identify the steering direction.}
%  \label{fig:act_map}
% %  \vspace{0.5cm}
% \end{figure}

\begin{figure}[!h]
   \centering
\resizebox{\linewidth}{!}{
\setlength{\tabcolsep}{2pt}
\begin{tabular}{ccc}

\shortstack{\includegraphics[width=0.33\linewidth]{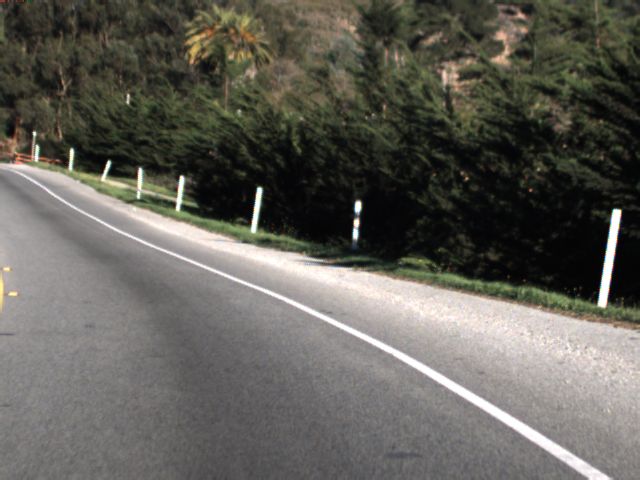}}&
\shortstack{\includegraphics[width=0.33\linewidth]{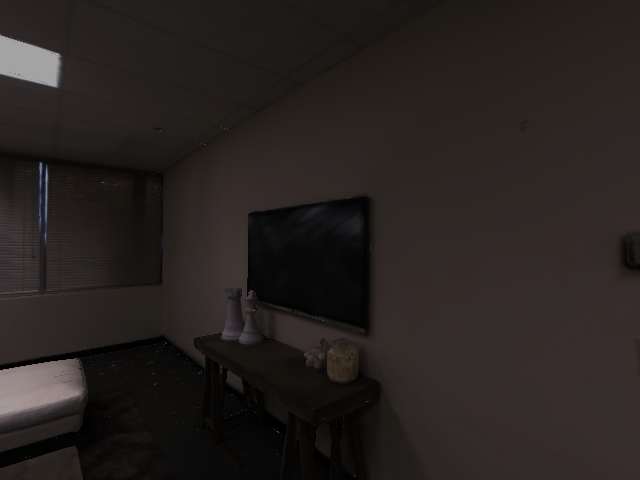}}&
\shortstack{\includegraphics[width=0.33\linewidth]{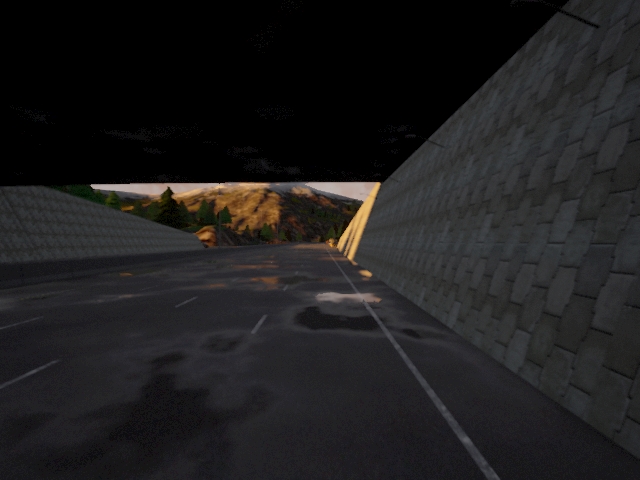}}\\[1pt]
% \hline\\
\shortstack{\includegraphics[width=0.33\linewidth]{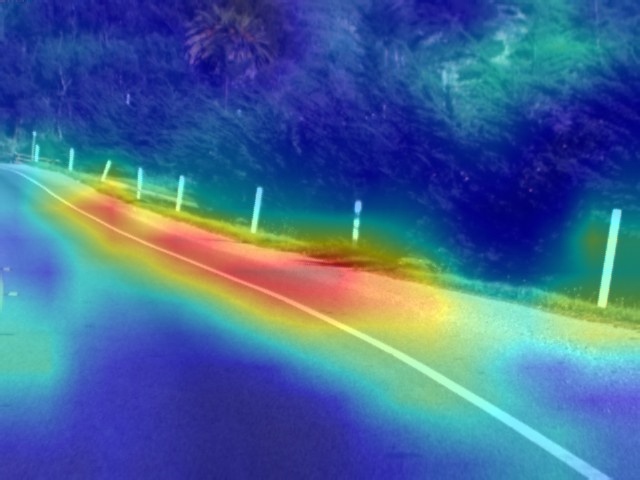} \\ (a) Udacity}&
\shortstack{\includegraphics[width=0.33\linewidth]{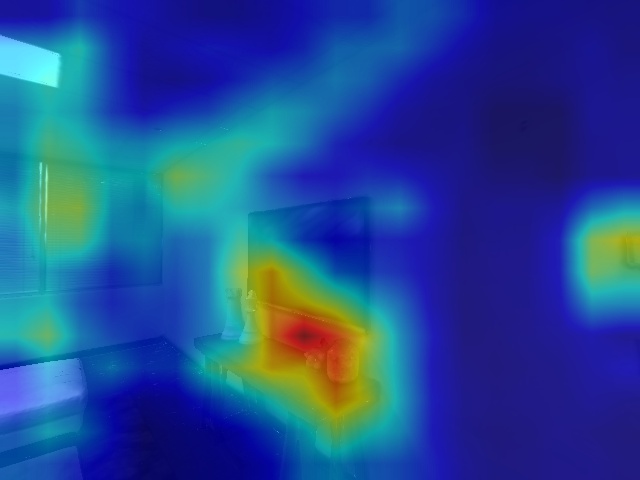} \\ (b) Gazebo}&
\shortstack{\includegraphics[width=0.33\linewidth]{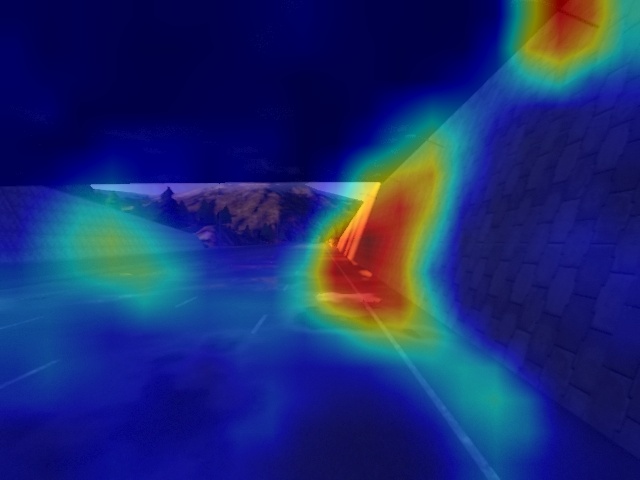} \\ (c) Carla}\\ [1pt]
\end{tabular}
}
%\vspace{3ex}
    \caption{Spatial support regions for predicting steering angle in three datasets. In most cases, we can observe that our FADNet focuses on ``line-like” patterns to predict the driving direction.
    }
    \label{fig:act_map}
\end{figure}

\subsection{Ablation Study}
\label{eff_def}

\begin{table}[h]
\begin{center}
%\footnotesize
\small
% \vspace{+0.15 cm}
\setlength{\tabcolsep}{0.66 em} % for the horizontal padding
{\renewcommand{\arraystretch}{1.2}% for the vertical padding
\begin{tabular}{|c|c|c|c|c|}
\hline
\multirow{2}{*}{\textbf{Architecture}} & \multirow{2}{*}{\textbf{\begin{tabular}[c]{@{}c@{}}Learning\\ Method\end{tabular}}} & \multicolumn{3}{c|}{\textbf{Dataset}}                                          \\ \cline{3-5} 
                                       &                                                                                     & \textit{\textbf{Udacity}} & \textit{\textbf{Gazebo}} & \textit{\textbf{Carla}} \\ \hline
\multirow{3}{*}{DroNet~\cite{loquercio2018dronet}}                & CLL~\cite{loquercio2018dronet}                                                                                 & 0.110                     & 0.082                    & 0.333                   \\ \cline{2-5} 
                                       & SFL~\cite{sattler2019robust}                                                                                 & 0.176                     & 0.081                    & 0.297                   \\ \cline{2-5} 
                                       & DFL (ours)                                                                          & 0.152                     & 0.073                    & 0.244                   \\ \hline
\multirow{3}{*}{FADNet (ours)}           & CLL~\cite{loquercio2018dronet}                                                                                   & 0.142                     & 0.081                    & 0.303                   \\ \cline{2-5} 
                                       & SFL~\cite{sattler2019robust}                                                                                 & 0.151                     & 0.071                    & 0.211                   \\ \cline{2-5} 
                                       & DFL (ours)                                                                          & \textbf{0.107}            & \textbf{0.069}           & \textbf{0.203}          \\ \hline
\end{tabular}
}
\end{center}
% % \vspace{-0.5cm}
\caption{Performance of different learning methods.
}
\label{tab:abl_DFSA}
\end{table}

\textbf{Effectiveness of our DFL.}
Table~\ref{tab:abl_DFSA} summarises the accuracy of DroNet~\cite{loquercio2018dronet} and our FADNet when we train them using different learning methods: CLL, SFL, and our peer-to-peer DFL. From this table, we can see that training both DroNet and FADNet with our peer-to-peer DFL clearly improves the accuracy compared with the SFL approach. This confirms the robustness of our fully decentralized approach and removes a need of a central server when we train a deep network with FL. Compared with the traditional CLL approach, our DFL also shows a competitive performance. However, we note that training a deep architecture using CLL is less complicated than with SFL or DFL. Furthermore, CLL is not a federated learning approach and does not take into account the privacy of the user data.

\textbf{Effectiveness of our FADNet.} Table~\ref{tab:abl_DFSA} shows that apart from the learning method, the deep architectures also affect the final results. This table illustrates that our FADNet combined with DFL outperforms DroNet in all configurations. We notice that DroNet achieves competitive results when being trained with CLL. However DroNet is not designed for federated training, hence it does not achieve good accuracy when being trained with SFL or DFL. On the other hand, our introduced FADNet is particularly designed with dedicated layers to handle the data imbalance and model convergence problem in federated training. Therefore, FADNet achieves new state-of-the-art results in all three datasets.

\textbf{Network Topology Analysis.}
Table~\ref{tab:cross_silo} illustrates the performance of DroNet and our FADNet when we train them using DFL under two distributed network topologies: Gaia and NWS. This table shows that the results of DroNet and FADNet under DFL are stable in both Gaia and NWS distributed networks. We note that the NWS topology has 22-silos while the Gaia topology has only 11 silos. This result validates that our FADNet and DFL do not depend on the distributed network topology. Therefore, we can potentially use them in practice with more silo data.

\begin{table}[]
\begin{center}
%\footnotesize
\small
% \vspace{+0.15 cm}
\setlength{\tabcolsep}{0.55 em} % for the horizontal padding
{\renewcommand{\arraystretch}{1.2}% for the vertical padding
\begin{tabular}{|c|c|c|c|c|}
\hline
\multirow{2}{*}{\textbf{\begin{tabular}[c]{@{}c@{}}Network\\Topology\end{tabular}}}                                          & \multirow{2}{*}{\textbf{Architecture}} & \multicolumn{3}{c|}{\textbf{Dataset}}                                     \\ \cline{3-5} 
                                                                           &                                        & \textit{\textbf{Udacity}} & \textit{\textbf{Gazebo}} & \textit{\textbf{Carla}} \\ \hline
\multirow{2}{*}{\begin{tabular}[c]{@{}c@{}}Gaia\\ (11 silos)\end{tabular}} & DroNet~\cite{loquercio2018dronet}                                 & 0.152                     & 0.073                 & 0.244                 \\ \cline{2-5} 
                                                                           & FADNet (ours)                                   & 0.107                     & 0.069                 & 0.203                 \\ \hline
\multirow{2}{*}{\begin{tabular}[c]{@{}c@{}}NWS\\ (22 silos)\end{tabular}}  & DroNet~\cite{loquercio2018dronet}                                 & 0.157                     & 0.075                 & 0.239                 \\ \cline{2-5} 
                                                                           & FADNet (ours)                                   & 0.109                     & 0.070                 & 0.200                 \\ \hline
\end{tabular}
}
\end{center}
\caption{Performance in different network topologies.}
\label{tab:cross_silo}
\vspace{-2ex}

\end{table}

\begin{figure}[!t]
   \centering
   \subfigure[Gazebo]{\includegraphics[width=0.49\linewidth]{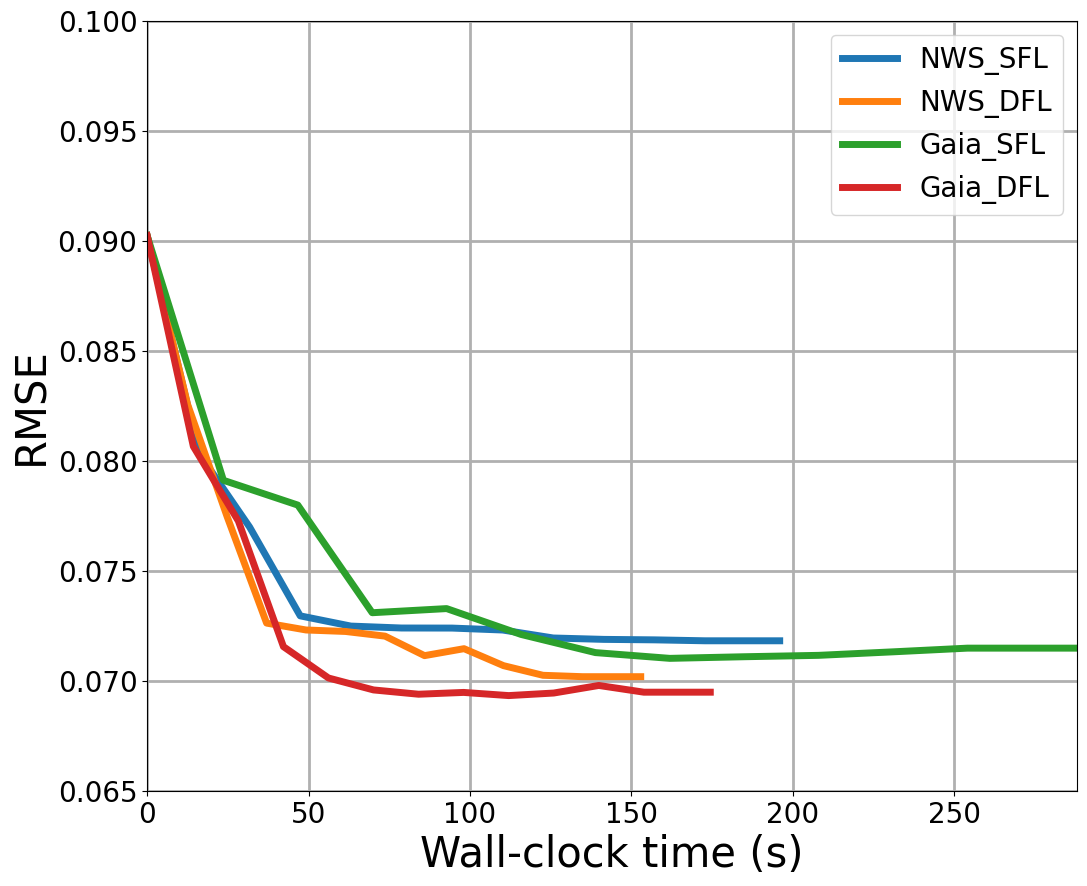}}
   \subfigure[Carla]{\includegraphics[width=0.49\linewidth]{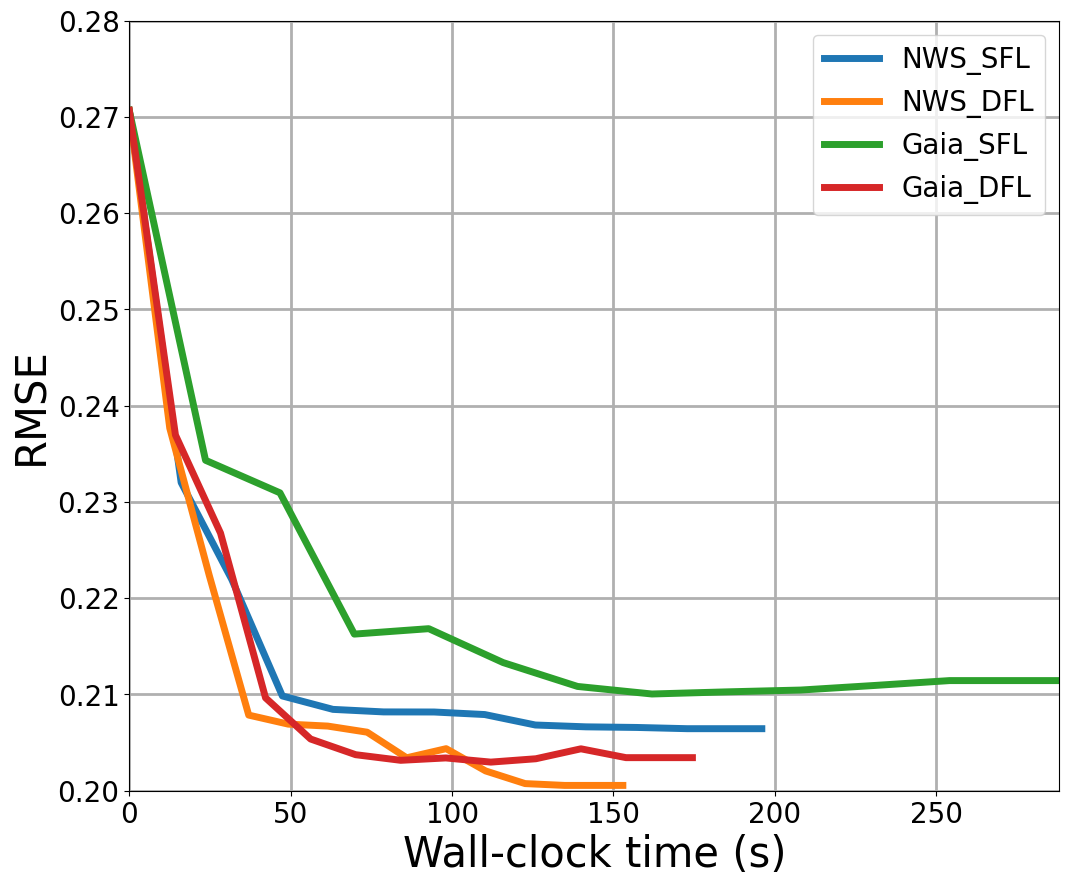}}
   \vspace{-1ex}
 \caption{The convergence ability of our FADNet and DFL under Gaia and NWS topology. Wall-clock time or elapsed real-time is the actual time taken from the start of the whole training process to the end, including the synchronization time of the weight aggregation process. All experiments are conducted with $3,000$ communication rounds.
 }
 \label{fig:graph}
  \vspace{-2ex}
\end{figure}

\textbf{Convergence Analysis.} The effectiveness of federated learning algorithms is identified through the convergence ability, including accuracy and training speed, especially when dealing with the increasing number of silos in practice. Fig.~\ref{fig:graph} shows the convergence ability of our FADNet with DFL using two topologies: Gaia~\cite{knight2011internetzoo} with 11 silos, and NWS~\cite{awscloud} with 22 silos. This figure shows that our proposed DFL achieves the best results in Gaia and NWS topology and converges faster than the SFL approach in both Gazebo and Carla datasets. We also notice that the performance of our DFL is stable when there is an increase in the number of silos. Specifically, training our FADNet with DFL reaches the converged point after approximately $150$s, $180$s on the NWS and Gaia topology, respectively. Fig.~\ref{fig:graph} validates the convergence ability of our FADNet and DFL, especially when dealing with the increasing number of silos.

In practice, compared with the traditional CLL approach, federated learning methods such as SFL or DFL can leverage more GPUs remotely. Therefore, we can reduce the total training time significantly. However, the drawback of federated learning is we would need more GPUs in total (ideally one for each silo), and deep architecture also should be carefully designed to ensure model convergence.

%%%%%%%%%%%%%%%%%%%%%%%%%%%%%%%%%%%% 
% WILL COPY TO INTRO
%%%%%%%%%%%%%%%%%%%%%%%%%%%%%%%%%%%% 
% \textbf{Tuong: I write the disadvantage of distributed learning FL/DFL here in case you need it.}

% Two main disadvantage of distributed learning :
% \textbf{Non I.I.D}: input images and output distributions in different silos are different; no data sharing between silos. This is one of the most critical problems of distributed learning since it can make the model not be converged. The solution for this problem may come from model architecture design and the distributed learning method.

% \textbf{Internet Connection:} Unlike local training, synchronization in distributed learning takes time. It can be a trade-off with the training time of the model in each silo. If the time spent for synchronization is longer than the training time, the efficiency of distributed learning is then limited.

\subsection{Deployment}
To verify the effectiveness of our FADNet in practice, we deploy the model trained on the Gazebo dataset on a mobile robot. The robot is equipped with a RealSense camera to capture the front RGB images. Our FADNet is deployed on a Qualcomm RB5 board to make the prediction of the steering angle for the robot. The processing time of our FADNet on the Qualcomm RB5 board is approximately $12$ frames per second. Overall, we observe that the robot can navigate smoothly in an indoor environment without colliding with obstacles. More qualitative results can be found in our supplementary material.

\section{Conclusion}
We propose a new approach to learn an autonomous driving policy from sensory data without violating the user's privacy. We introduce a peer-to-peer deep federated learning (DFL) method that effectively utilizes the user data in a fully distributed manner. Furthermore, we develop a new deep architecture - FADNet that is well suitable for distributed training. The intensive experimental results on three datasets show that our FADNet with DFL outperforms recent state-of-the-art methods by a fair margin. Currently, our deployment experiment is limited to a mobile robot in an indoor environment. In the future, we would like to test our approach with more silos and deploy the trained model using an autonomous car on man-made roads.

\bibliographystyle{class/IEEEtran}
\bibliography{class/reference}
%   class/IEEEabrv,
\end{document}